\documentclass[final]{cvpr}

\usepackage{times}
\usepackage{epsfig}
\usepackage{graphicx}
\usepackage{amsmath}
\usepackage{amssymb}

\usepackage{booktabs}
\usepackage{multirow}
\usepackage{subfigure}
\usepackage{float}

\usepackage[xindy,acronym]{glossaries}
\glsdisablehyper
\defglsdisplayfirst[acronym]{\emph{#1#4}}

\newacronym{gl:DSM}{DSM}{digital surface model}
\newacronym{gl:nDSM}{nDSM}{normalized digital surface model}
\newacronym{gl:PAN}{PAN}{panchromatic}
\newacronym{gl:cGAN}{cGAN}{conditional generative adversarial network}
\newacronym{gl:CBAM}{CBAM}{convolutional block attention module}
\newacronym{gl:cLSGAN}{cLSGAN}{convolutional least squares GAN}
\newacronym{gl:NMS}{NMS}{non-maximum suppression}
\newacronym{gl:CCL}{CCL}{connected component labeling}
\newacronym{gl:DFS}{DFS}{depth first search}
\newacronym{gl:CityGML}{CityGML}{city geography markup language}
\newacronym{gl:MAE}{MAE}{mean absolute error}
\newacronym{gl:RMSE}{RMSE}{root mean squared error}
\newacronym{gl:NMAD}{NMAD}{normalised median absolute deviation}
\newacronym{gl:IoU}{IoU}{intersection over union}
\newacronym{gl:GAN}{GAN}{generative adversarial network}\newacronym{gl:LoD}{LoD}{level of detail}
\newacronym{gl:LiDAR}{LiDAR}{light detection and ranging}
\newacronym{gl:PCA}{PCA}{principle component analysis}
\newacronym{gl:FFT}{FFT}{fast Fourier transform}

\usepackage[pagebackref=true,breaklinks=true,colorlinks,bookmarks=false]{hyperref}
\usepackage[capitalise,noabbrev]{cleveref}

\begin{document}

\AddToShipoutPicture*{%
\AtTextUpperLeft{%
\put(0,\LenToUnit{1cm}){%
\fbox{\parbox{\textwidth}{%
\centering
This work was accepted to be presented at the IEEE/GRSS Workshop on Large Scale Computer Vision for Remote Sensing Imagery (EarthVision) to be held at the IEEE Conference on Computer Vision and Pattern Recognition (CVPR) 2021.
}}%
}%
}%
}%

\title{Machine-learned 3D Building Vectorization from Satellite Imagery}


\author{Yi Wang$^1$, Stefano Zorzi$^2$, Ksenia Bittner$^1$\\

$^1$German Aerospace Center (DLR), $^2$Graz University of Technology\\

{\tt\small \{yi.wang,ksenia.bittner\}@dlr.de}, {\tt\small stefano.zorzi@icg.tugraz.at}
}

\maketitle

\begin{abstract}
We propose a machine learning based approach for automatic 3D building reconstruction and vectorization. Taking a single-channel photogrammetric \gls{gl:DSM} and a \gls{gl:PAN} image as input, we first filter out non-building objects and refine the building shapes of the input \gls{gl:DSM} with a \gls{gl:cGAN}. The refined \gls{gl:DSM} and the input \gls{gl:PAN} image are then used through a semantic segmentation network to detect edges and corners of building roofs. Later, a set of vectorization algorithms are proposed to build roof polygons. Finally, the height information from refined \gls{gl:DSM} is processed and added to the polygons to obtain a fully vectorized \gls{gl:LoD}-2 building model. We verify the effectiveness of our method on large-scale satellite images, where we obtain state-of-the-art performance.
\end{abstract}

\glsresetall
\section{Introduction}
The availability of accurate 3D building models has become highly demanded in various applications like the modeling of global urbanization process, urban planning, disaster monitoring, \etc. As traditional methods performed by human operators for 3D building modeling are expensive, time-consuming and limited to a small area, modern automatic 3D building model reconstruction methods have drawn wide research interests.

Current automatic 3D building reconstruction methods can be generally categorized into data-driven, model-driven and hybrid approaches. While model-driven approaches extract the primitives of buildings and fit them to the most appropriate models~\cite{lafarge2008structural}, data-driven methods extract geometrical components of building roof planes from 3D point clouds or \glspl{gl:DSM} with point- or image-based segmentation techniques, and these components are merged to 3D models with respect to some geometrical topology~\cite{tarsha2007model}. With model-driven methods being unable to solve complex situations and data-driven methods being commonly noisy, hybrid approaches, including this work, tend to integrate the two types of approaches, where a data-driven approach extracts the building components, and a model-driven approach utilizes prior knowledge of the geometrical building models to help reconstruct 3D buildings~\cite{zheng2017hybrid}. 


\begin{figure}[t]
\begin{center}
\subfigure[]{
\includegraphics[width=0.25\linewidth]{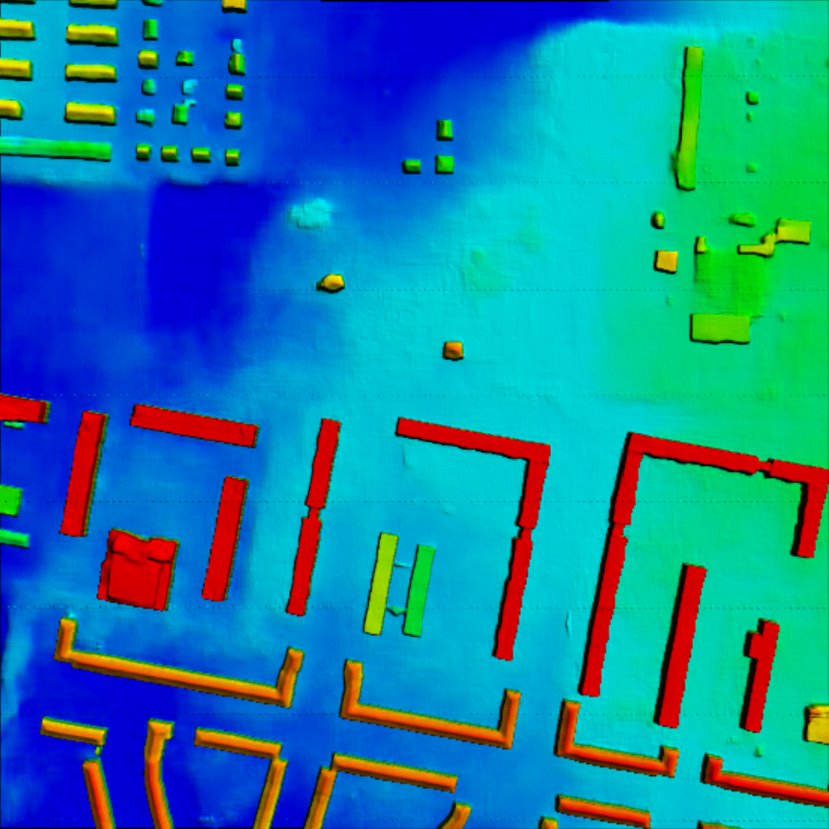}
}
\subfigure[]{
\includegraphics[width=0.25\linewidth]{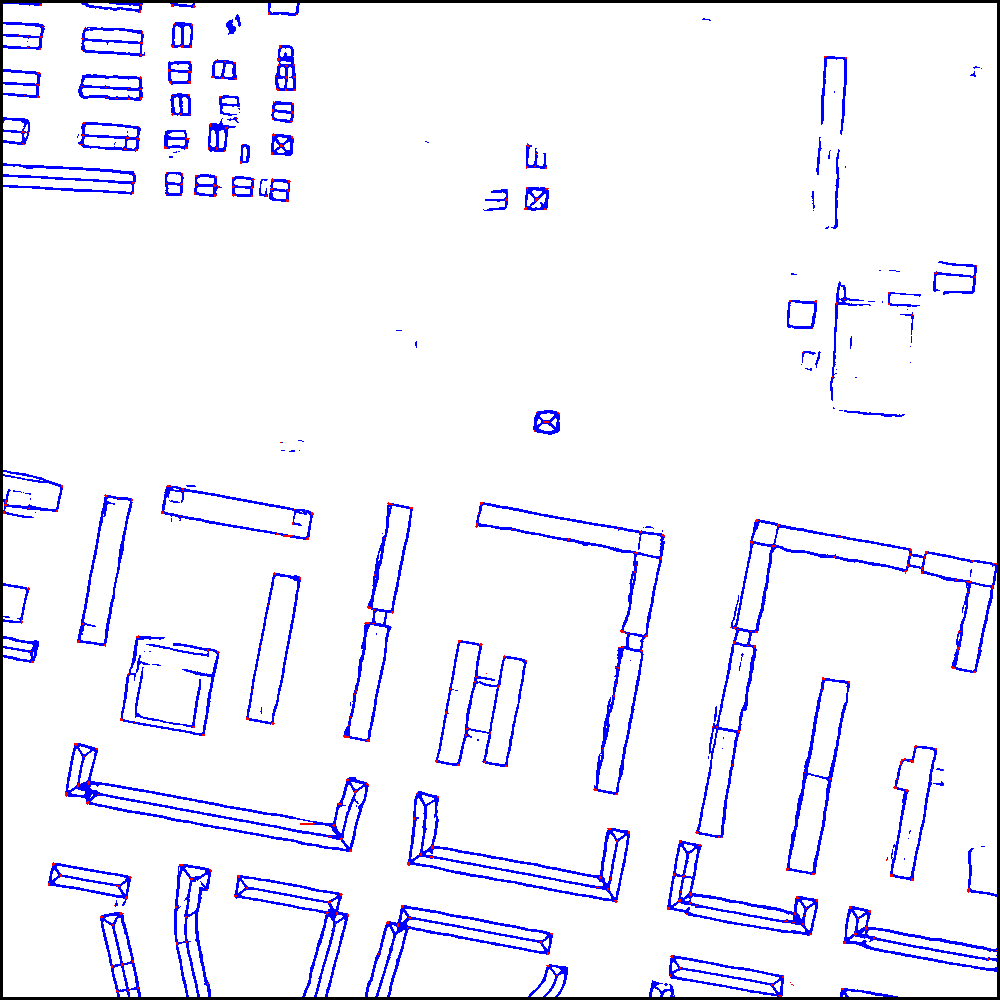}
}
\subfigure[]{
\includegraphics[width=0.4\linewidth]{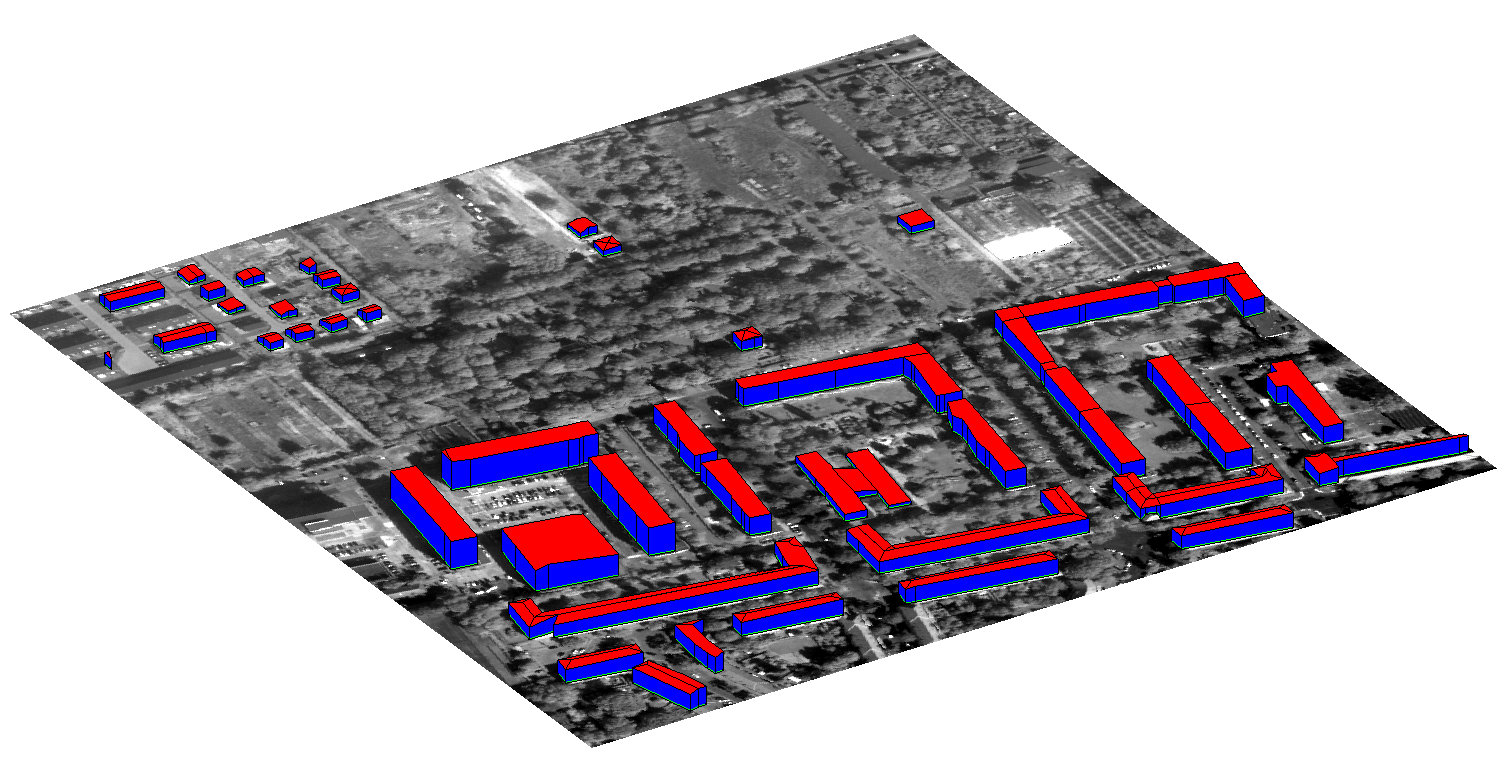}
}
\end{center}
   \caption{Sample results of the proposed 3D building vectorization method. $(a)$ refined DSM; $(b)$ edge and corner segmentation; $(c)$ vectorized 3D building model.}
\label{fig:small_overview}
\end{figure}

While \gls{gl:LiDAR} point clouds and aerial images have been the most common sources to extract 3D building information in the past years~\cite{brenner2005building,chen2005building,haala2010update}, satellite images become more and more important as they are convenient to acquire, cover wide areas and update frequently. Apart from optical images, modern satellites can also provide \glspl{gl:DSM} using photogrammetric stereo matching techniques, from which we can extract both building objects and their height information. However, satellite \glspl{gl:DSM} show a reasonable amount of noise and outliers because of matching errors or the existence of non-building objects, thus refinement methods have been studied to improve their quality. With traditional methods using filter-based techniques like \gls{gl:PCA}~\cite{lopez2000improving}, Kalman filter~\cite{wang1998applying} and \gls{gl:FFT}~\cite{arrell2008spectral} to remove outliers, recent researches have shown promising improvement by using deep learning based methods. Bittner \etal~\cite{bittner2018automatic} proposed firstly a \gls{gl:cGAN} based approach to filter out non-building objects and refine building shapes of photogrammetric \glspl{gl:DSM}, which was further developed by a set of works~\cite{bittner2019multi,bittner2020long,bittner2019late} to step-by-step improve the generation quality. Stucker and Schindler~\cite{stucker2020resdepth} proposed an improvement for traditional stereo image matching by regressing a residual correction with a convolutional neural network.

The revolutionary appearance of machine learning and deep learning techniques has also brought significant contributions to the whole process of 3D building reconstruction tasks. Not only building footprints can be extracted and regularized with neural networks~\cite{vakalopoulou2015building,zhao2018building,zorzi2020machine}, but also the heights and roof elements can be detected and predicted~\cite{alidoost2020shaped,alidoost20192d}, leading to constructed 3D building models. Recent researches can be found in~\cite{mahmud2020boundary}, where the authors combined building object detection, semantic segmentation and height prediction in a multi-task manner, and~\cite{wang2017single}, where the authors proposed a deep learning based model-driven approach to perform parametric building reconstruction. While most of these researches focusing on \gls{gl:LoD}-1, \gls{gl:LoD}-2 building modeling is relatively new. One example is presented in~\cite{partovi2019automatic}, where a hybrid 3D building reconstruction method is applied to detect and decompose building boundaries, classify roof types, and fit predefined building models.

Challenges for \gls{gl:LoD}-2 building reconstruction contain the requirement for accurate building height prediction and roof element extraction, and the complexity to form vectorized 3D roofs. Most existing methods utilize or predict coarse height maps for detection tasks of neural networks and later perform optimization~\cite{alidoost20192d,partovi2019automatic}. Our work, by contrast, uses network refined \glspl{gl:DSM} to extract roof elements and proposes a corresponding vectorization pipeline to form 3D models.

In this paper, we propose a machine learning based approach to reconstruct \gls{gl:LoD}-2 building models from photogrammetric \glspl{gl:DSM} and \glspl{gl:PAN} image obtained from satellites. Our contributions can be described as following:

\begin{itemize}
    \item We improve the state-of-the-art \gls{gl:cGAN} based \gls{gl:DSM} refinement network proposed by Bittner \etal~\cite{bittner2020long} by adding a popular self-attention \gls{gl:CBAM}~\cite{woo2018cbam}.
    \item We propose an edge and corner detection network sharing the architecture of the previous \gls{gl:DSM} refinement network.
    \item We propose a novel vectorization pipeline to polygonize building roofs and reconstruct 3D building models.
\end{itemize}

\section{Methodology}
\begin{figure*}
\centering
\includegraphics[width=\textwidth]{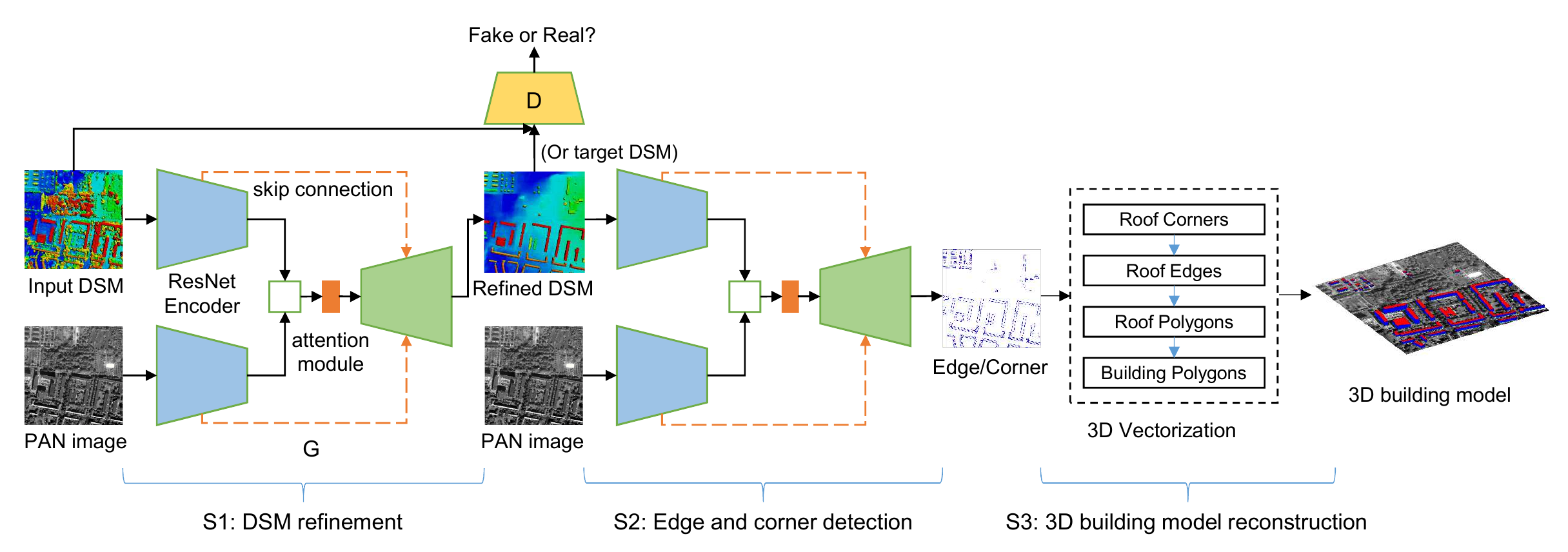}
\caption[overview]{Overview of the proposed method. Given a photogrammetric \gls{gl:DSM} and a \gls{gl:PAN} image as input, a \gls{gl:cGAN} based DSM refinement network and a semantic segmentation network are sequentially applied to refine building shapes and detect edges and corners. A set of vectorization algorithms are then applied to reconstruct a full 3D building model.}
\label{fig:overview}
\end{figure*}

As is shown in \cref{fig:overview}, our multi-stage 3D building vectorization approach starts with a \gls{gl:cGAN} architecture for photogrammetric \gls{gl:DSM} building shape refinement. The refined \gls{gl:DSM}, together with the input PAN image, is then used to detect building edges and corners with a semantic segmentation network that shares the structure of the cGAN generator. The detected edges and corners are later vectorized to building roof polygons. In the final stage, the refined \gls{gl:DSM} and 2D polygons are combined to reconstruct 3D building models.

\subsection{DSM building shape refinement}

\begin{figure}[ht]
\begin{center}
\includegraphics[width=\linewidth]{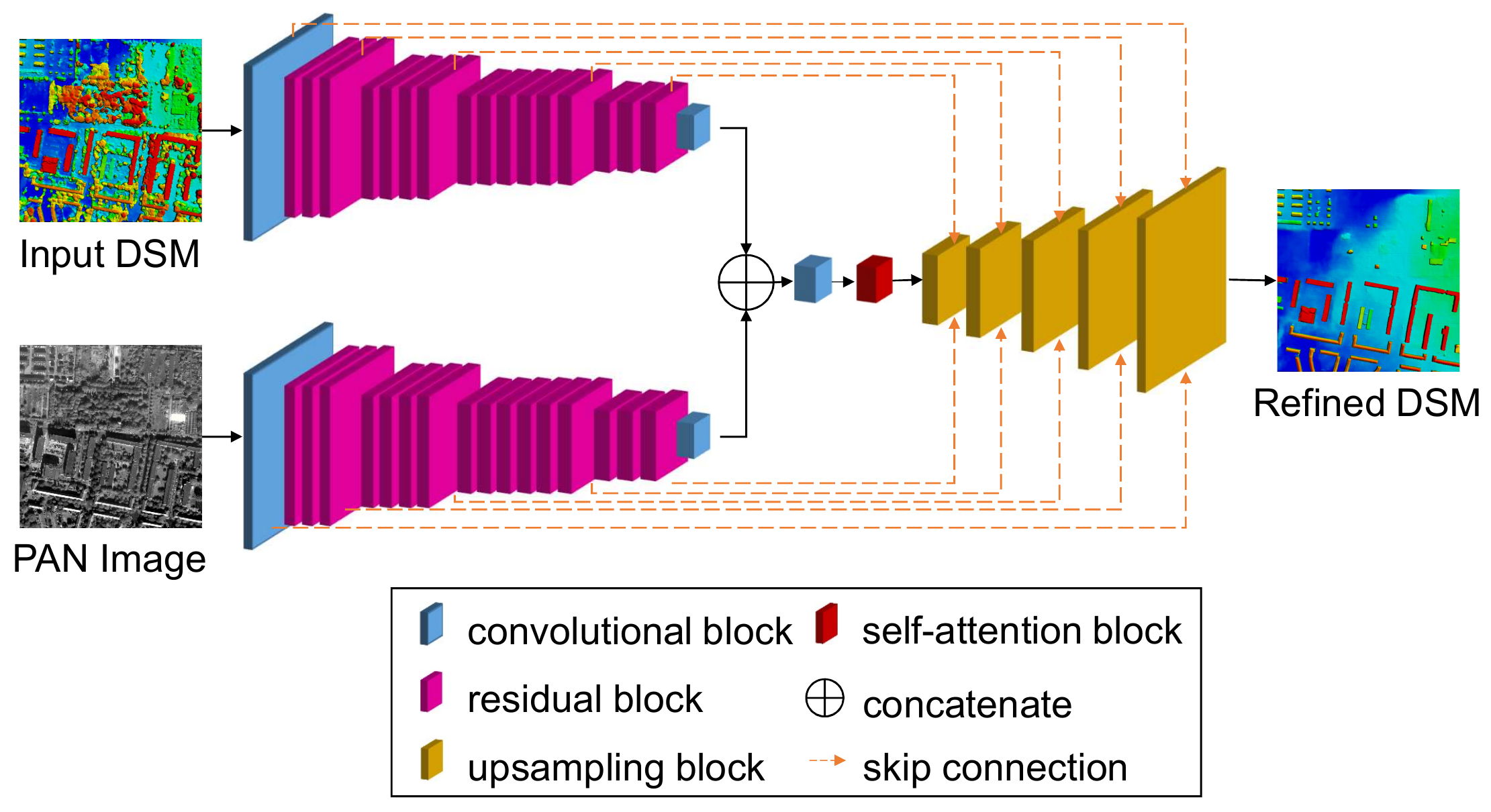}
\end{center}
   \caption{Generator architecture of the proposed \gls{gl:DSM} refinement network.}
\label{fig:generator}
\end{figure}

The proposed deep neural network for \gls{gl:DSM} refinement is an extension of the network presented by Bittner \etal~\cite{bittner2020long} based on an image-to-image translation \gls{gl:cGAN} introduced by Isola \etal~\cite{isola2017image}. The network jointly learns a generator and a discriminator to do the domain transfer, \ie from a source domain, the photogrammetric \gls{gl:DSM}, to a target domain, the refined \gls{gl:DSM}. With the discriminator following the PatchGAN architecture proposed by Isola \etal~\cite{isola2017image}, the generator has a UNet-like structure with both long skip connections from the encoders to the decoder and short skip connections in-between the residual blocks inside the encoders. To enhance the feature of building objects, we add a \gls{gl:CBAM} as presented by Woo \etal~\cite{woo2018cbam} before the decoder. The \gls{gl:CBAM} is a combination of 1D channel attention and 2D spatial attention, which are sequentially multiplied to the input feature maps. The overall generator architecture is shown in~\cref{fig:generator}.

Following the idea presented by Bittner \etal~\cite{bittner2020long}, we combine several types of losses in a multi-task manner for optimizing the proposed \gls{gl:DSM} refinement network:

\begin{equation}
\label{total_loss}
\begin{array}{c}
\mathcal{L}_{\text {total}}\left(G\right) = \alpha \cdot \mathcal{L}_{\mathrm{cLSGAN}}(G,D)+\beta \cdot \mathcal{L}_{L_{1}}(G)\\

+ \gamma \cdot \mathcal{L}_{\text {normal }}\left(\mathcal{N}^{\mathrm{t}}, \mathcal{N}^{\mathrm{p}}\right)
\end{array}
\end{equation}

\noindent
where $\alpha$, $\beta$ and $\gamma$ represent the weighting parameters of different loss terms.

\vspace{1em}
\noindent \textbf{GAN loss.}
We combine a conditional GAN~\cite{mirza2014conditional} and a Least Squares GAN~\cite{mao2017least} for the \gls{gl:DSM} refinement network, thus a \gls{gl:cLSGAN} loss is utilized:

\begin{equation}
\begin{array}{c}
\underset{G}{\min} \underset{D}{\max} \mathcal{L}_{\text {cLSGAN }}(G, D)=\mathbb{E}_{x, y \sim p_{\text {real }}(y)}\left[(D(y,x)-1)^{2}\right] \\

+\mathbb{E}_{x, z \sim p_{z}(z)}\left[D(G(z,x),x)^{2}\right]
\end{array}
\end{equation}

\noindent
where $y \sim p_{\text {real }}(y)$ represents real samples, and $G(z)$ represents generated samples transferred from usually latent noise variables $z \sim p_{z}(z)$. Respectively, $x$ denotes the \gls{gl:GAN}'s condition (the input \gls{gl:DSM}), $D(y,x)$ represents discriminator output of real samples, and $D(G(z,x),x)$ represents discriminator output of generated samples. 

\vspace{1em}
\noindent \textbf{L1 loss.}
It is common to blend the objective functions for \glspl{gl:GAN} with traditional regression losses like $L1$ or $L2$ distances to help the generator create images as close as possible to the given ground truth. Since $L1$ loss encourages less blurring to the image boundaries, it is added to our generator losses:

\begin{equation}
\mathcal{L}_{L_{1}}(G)=\mathbb{E}_{x, y \sim p_{\text {real }}(y), z \sim p_{z}(z)}\left[\|\mathrm{y}-G(z,x)\|_{1}\right]
\end{equation}

\vspace{1em}
\noindent \textbf{Normal vector loss.}
To further refine the surface of building roof planes, a normal vector loss~\cite{hu2019revisiting}, which measures the angles between normal vectors of generated and target \glspl{gl:DSM}, is added to the generator losses:

\begin{equation}
\mathcal{L}_{\text {normal }}\left(\mathcal{N}^{\mathrm{t}}, \mathcal{N}^{\mathrm{p}}\right)=\frac{1}{m} \sum_{i=1}^{m}\left(1-\frac{\left\langle n_{i}^{\mathrm{t}}, n_{i}^{\mathrm{p}}\right\rangle}{\left\|n_{i}^{\mathrm{t}}\right\|\left\|n_{i}^{\mathrm{p}}\right\|}\right),
\end{equation}

\noindent
where $\mathcal{N}^{\mathrm{t}}=\left\{n_{1}^{\mathrm{t}}, \ldots, n_{m}^{\mathrm{t}}\right\}$ and $\mathcal{N}^{\mathrm{p}}=\left\{n_{1}^{\mathrm{p}}, \ldots, n_{m}^{\mathrm{p}}\right\}$ represent normal vectors of the target and predicted \gls{gl:DSM}, and $\langle \cdot,\cdot \rangle$ denotes the scalar product of the two vectors. This normal vector loss emphasizes the planarity and inclination of building roofs. The smaller the angle, the more planar the predicted surface and the more consistent to the target surface.

The combination of different losses forms a multi-task learning problem, thus an automatic weighting method proposed firstly by Kendal \etal~\cite{kendall2018multi} and investigated in remote sensing in~\cite{liebel2020generalized,liebel2018auxiliary} is applied to automatically tune the loss weights considering the homoscedastic uncertainty of each separate task:

\begin{equation}
w_{l}=\left\{\begin{array}{ll}
0.5 \cdot \exp \left(-\log \left(\sigma_{l}^{2}\right)\right) & \text { for } \mathcal{L}_{L_{1}} \text { and } \mathcal{L}_{\text {normal }} \\
\exp \left(-\log \left(\sigma_{l}^{2}\right)\right) & \text { for } \mathcal{L}_{\text {cLSGAN }}
\end{array}\right.
\label{eq:weighting}
\end{equation}

\noindent
where $\sigma_{l}^{2}$ is a learnable parameter, which represents the variance, \ie uncertainty of each task through the training process. In order to avoid over-controlled parameter values, a regularization term $0.5 \cdot \log \left(\sigma_{l}^{2}\right)$ is added following each weighted loss. As a result, the final loss of the generator of this \gls{gl:DSM} refinement network can be formulated as:

\begin{equation}
\mathcal{L}_{\text {total}}\left(G\right)=\sum_{l} w_{l} \cdot \mathcal{L}_{l}+\mathcal{R}_{l}
\end{equation}

\noindent
while the discriminator loss remains the same as the \gls{gl:cLSGAN} loss:

\begin{equation}
\begin{array}{c}
\mathcal{L}_{\text {total }}(D)=\mathbb{E}_{x, y \sim p_{\text {real }}(y)}\left[(D(y,x)-1)^{2}\right] \\

+\mathbb{E}_{x, z \sim p_{z}(z)}\left[D(G(z,x),x)^{2}\right]
\end{array}
\end{equation}

\begin{figure*}[ht]
\centering
\includegraphics[width=\textwidth]{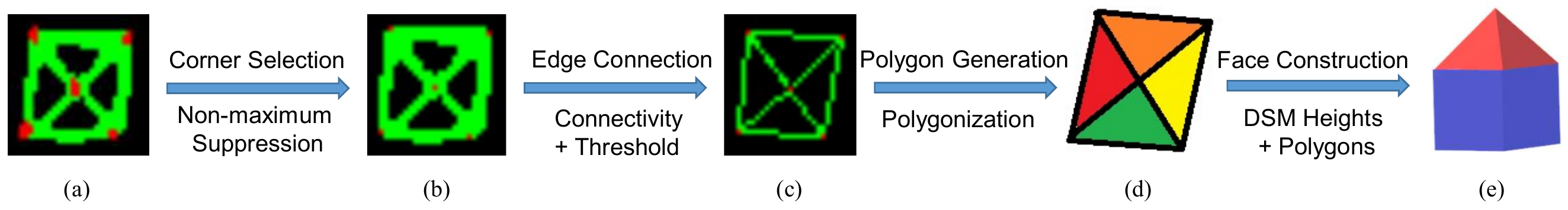}
\caption[vectorization]{Overview of the proposed vectorization pipeline. (a) detected edges and corners. (b) edges and filtered corners. (c) vectorized edges and corners. (d) roof polygons. (e) 3D building model.}
\label{fig:vectorization}
\end{figure*}

\subsection{Building edge and corner detection}

Given the refined \gls{gl:DSM} and PAN image, a semantic segmentation network is used to detect building edges and corners. The network architecture is identical to the generator of the \gls{gl:DSM} refinement network (see \cref{fig:generator}), except the change of the three-channel output layer. A simple multi-class cross-entropy loss is applied:

\begin{equation}
\mathcal{L}_{\mathrm{CE}}(x,t)=\mathbb{E}\left[-\sum_{i=1}^{3} t_{i} \log x_{i}\right]
\end{equation}

\noindent
where $x_{i}$ is the predicted probability for a certain class $i$, and $t_{i}$ is either 0 or 1 depending on the label of class $i$ for the corresponding target. The output probability remains for further processing.

\subsection{3D building model reconstruction}

\begin{figure}
\centering
\includegraphics[width=\linewidth]{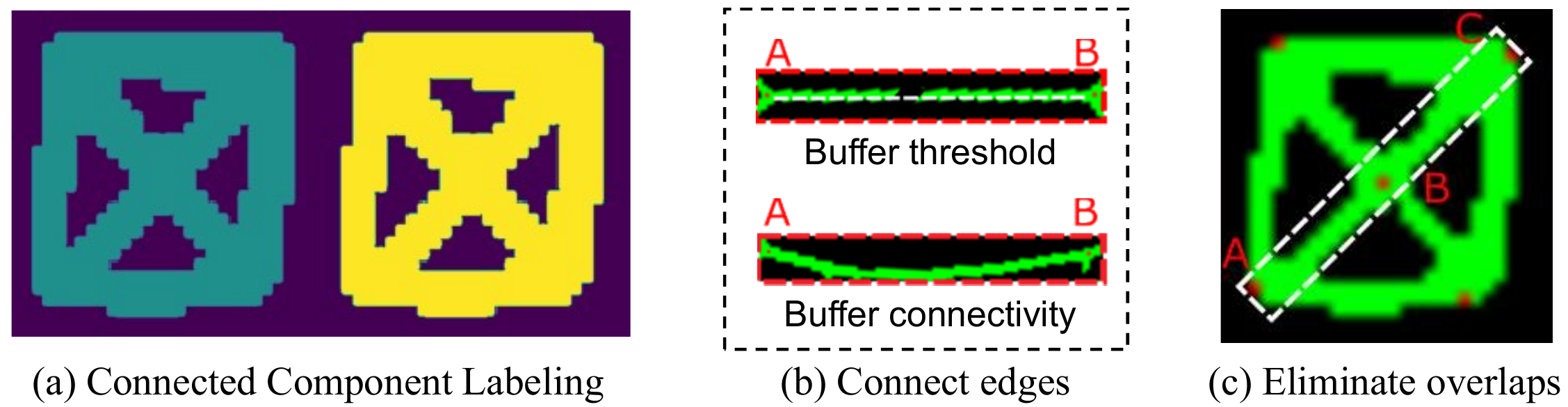}
\caption[building edge connection]{Examples of building edge vectorization.}
\label{fig:edge_vectorization}
\end{figure}

In the final stage, a novel 3D building vectorization method is proposed using the refined \gls{gl:DSM} and detected building edges and corners. Assuming building edges are straight lines, the core idea is to step-by-step build a graph data structure that stores points, lines, faces and their relationships for every single building. As being a hybrid method, the proposed approach is not limited to the complexity of different types of buildings, thus performing well especially for large area 3D building modeling. A general workflow is shown in \cref{fig:vectorization}.

\vspace{1em}
\noindent \textbf{Corner point selection.}
For each corner pixel in the ground truth, multiple surrounding pixels may be detected as corners, thus a \gls{gl:NMS} algorithm is implemented to filter out best fitting corner points. As is shown in \cref{fig:vectorization} $(a) - (b)$, for each detected corner pixel (the candidate), a surrounding $n \times n$ window is used as the evaluating box. For each neighbor pixel in this window, if the pixel value (corner probability) is no bigger than the candidate, it is set to zero; otherwise if it exceeds the candidate, this pixel remains while the candidate is set to zero. This process is iterated over all corner candidates and those isolated best candidates are seen as final corner points.


\vspace{1em}
\noindent \textbf{Roof edge vectorization.}
Before we start the vectorization process, a \gls{gl:CCL}~\cite{wu2005optimizing} algorithm is applied to label connected pixels into building instances. Two pixels are connected when they are neighbors and have a non-zero value. Here the neighborhood is defined in a 2-connected sense, which means every pixel has eight neighbors in eight directions. As shown in \cref{fig:edge_vectorization} $(a)$, different sets of connected pixels would have different IDs and separate different buildings, which enables the next steps to be performed within the scope of every single building.

Then we connect the corners to form edges based on two conditions. The first condition is the average pixel value of a line buffer between a pair of corner points. If the average value is above a threshold, an edge line is determined between the corners. This condition would possibly fail when the edge is curved in reality, thus a second condition is applied in parallel. By utilizing the \gls{gl:CCL} algorithm again in a rectangle buffer between the pair of corners, an edge is determined if the labels of the two corners are identical. Two examples are shown in \cref{fig:edge_vectorization} $(b)$, where both an edge with a hole and a little curved edge can be successfully detected.

With the two conditions we can efficiently and thoroughly detect building edges, yet still one problem needs to be considered. As it is shown in \cref{fig:edge_vectorization} $(c)$, corner $A$ and $B$, corner $B$ and $C$ form two edges, but corner $A$ and corner $C$ can also form an edge which is redundant since it covers $AB$ and $BC$. To solve this issue, we again create a rectangle buffer for each potential two-corner pair and, if other corner points exist inside this buffer, this pair can not form an edge anymore.

\vspace{1em}
\noindent \textbf{Roof polygon generation.}
The vectorized edges are then polygonized to roof faces ( see \cref{fig:vectorization} $(d)$), which can be easily done by graph search algorithms. For each building, an undirected graph is firstly built from the obtained edges. A simple \gls{gl:DFS} is then applied to detect and mark a cycle (\ie a roof polygon) in this graph by tracing a back edge to vertices that have been visited. This is run iteratively to extract all cycles with corresponding different marks. To avoid face overlapping, large cycles which cover small cycles are removed in the final step. In practice, the polygonization process can also be directly applied with a $polygonize$ function from the open-source \emph{shapely} package which is popular for manipulation and analysis of planar geometric objects~\cite{shapely}.

\begin{figure}
\centering
\includegraphics[width=0.8\linewidth]{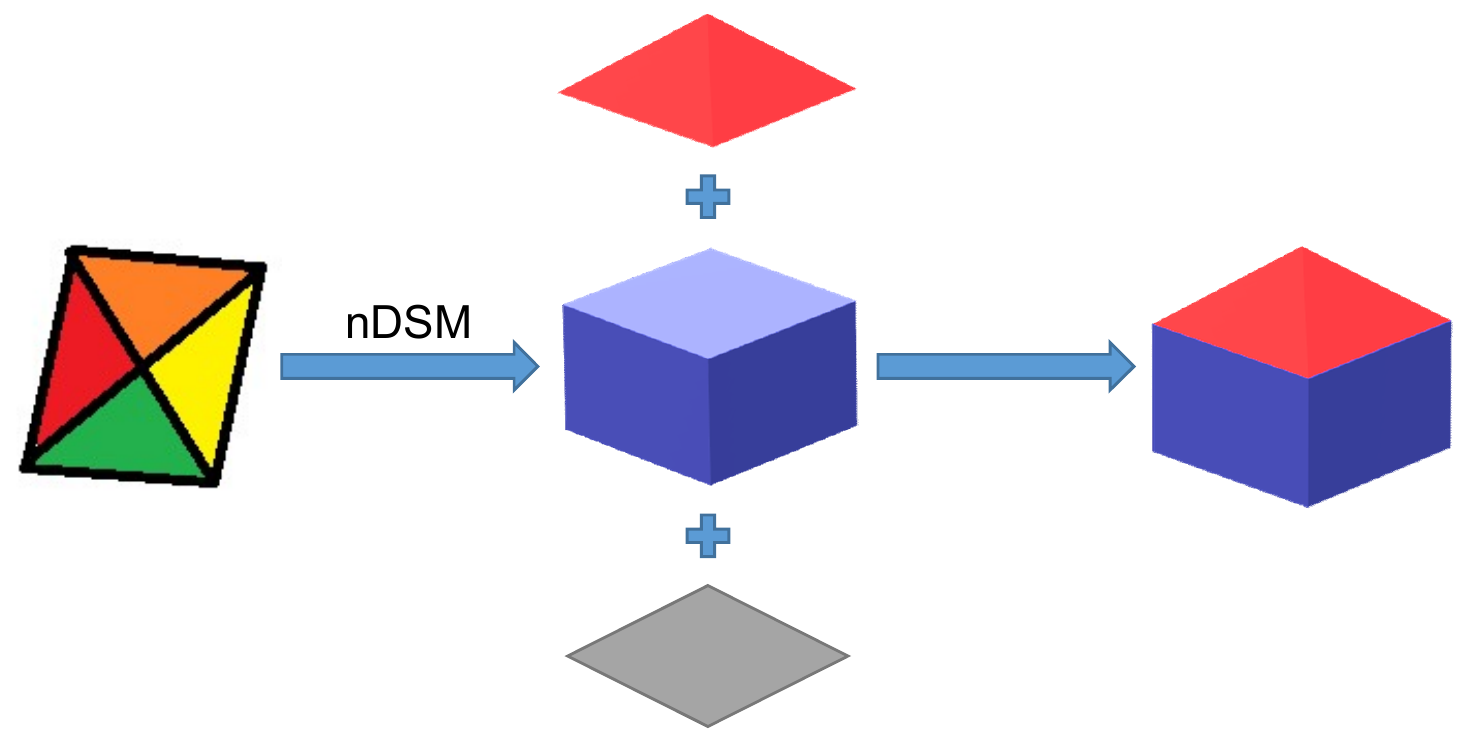}
\caption[constructing 3D building model]{The construction of final 3D building model. Height information from refined \gls{gl:DSM} is processed and added to the polygons to build 3D roofs, walls and ground face, together forming the final 3D model.}
\label{fig:3D modeling}
\end{figure}

\vspace{1em}
\noindent \textbf{3D building modeling.}
In the final stage, walls and the ground face are constructed utilizing roof polygons and the refined \gls{gl:DSM} to produce a full 3D building model. Firstly, a \gls{gl:nDSM} is generated from the refined \gls{gl:DSM} with the method proposed by Qin \etal~\cite{qin2016spatiotemporal}. Then the adjacent roof faces are merged into a union, \ie a polygon whose edges are the building outlines. This gives us the footprints of the building, which also means the 2D shape of the ground face. In the next step, the height information from the \gls{gl:nDSM} is applied to corner points both inside roofs and on the building boundaries. To avoid apparent height difference between endpoints of an edge due to corner miss-matching (especially on outer boundaries where corner height is supposed to be much bigger than neighboring ground pixels), a small window is applied again to adjust height values. This is done by giving the corner point the maximum height value in this small window. Though slightly decreasing general accuracy, it can largely improve the robustness and smoothness of resulting 3D models.

The edges of the union polygon represent both the upper and lower boundaries of the building's surrounding walls. With the height of upper corners already determined with the maximum height value in the window, the height of lower corners is determined by giving the minimum height value in the window, \ie zero, hence forming the building walls in 3D. Meanwhile, the lower edges form also the ground face of the building, resulting in the final 3D building model. The modeling process is shown in \cref{fig:3D modeling}.

\section{Experiments and results}

The proposed approach is evaluated on Worldview-1 data of Berlin, Germany. The input consists of a space-borne photogrammetric \gls{gl:DSM} and a panchromatic image with 0.5 $m$ spatial resolution covering a total area of 410 ${km}^2$. The ground truth is generated from the public \gls{gl:CityGML} dataset following the same procedure as described in~\cite{bittner2018dsm}. The \gls{gl:CityGML} data for Berlin is freely available at \url{https://www.businesslocationcenter.de/en/economic-atlas/download-portal/}. Open datasets for some other worldwide cities can be found at \url{https://3d.bk.tudelft.nl/opendata/opencities/}.

\subsection{Implementation details}

The \gls{gl:DSM} refinement network is based on the Coupled-UResNet \gls{gl:cGAN} architecture proposed by Bittner \etal \cite{bittner2020long}, with an additional \gls{gl:CBAM}~\cite{woo2018cbam} applied before the decoder. The edge and corner detection network shares the architecture of the generator of the \gls{gl:DSM} refinement network, while the last layer is changed to three-channel output with a \emph{$softmax$} activation function.

The networks are trained on a single NVIDIA TITAN X (PASCAL) GPU with 12 GB memory. To fit the training data into the GPU memory, the satellite images are tiled into 21480 samples of size 256$\times$256 px. A minibatch of 4 is applied in both networks. The samples are augmented not only by horizontal and vertical flipping but also tiled from the original image with a random overlap every epoch to give the model a clue about building geometries which happened to be on the patch border in previous epochs. During the training of both networks, the ADAM optimizer is used with an initial learning rate of $\alpha=0.0002$ and momentum parameters ${\beta}_1=0.5, {\beta}_2=0.999$. For the \gls{gl:DSM} refinement network, the generator is pre-trained for 100 epochs as a warm-up and later interpolated with the cGAN's generator. This so-called network interpolation~\cite{wang2018esrgan} can balance CNN's over-smoothing and GAN's over-sharpening. The initial learnable weighting parameters as described in \cref{eq:weighting} are equally set to 1.

During the vectorization process, the window size for both corner point filtering and corner height valuing is set to $5 \times 5$ pix, while the width for rectangle buffers (edge connecting and overlap elimination shown in \cref{fig:edge_vectorization} $(c)$) is set to 7 pix.

\subsection{Results and evaluation}


\begin{figure*}[htp]
\centering
\subfigure[PAN image]{\includegraphics[width=0.3\textwidth]{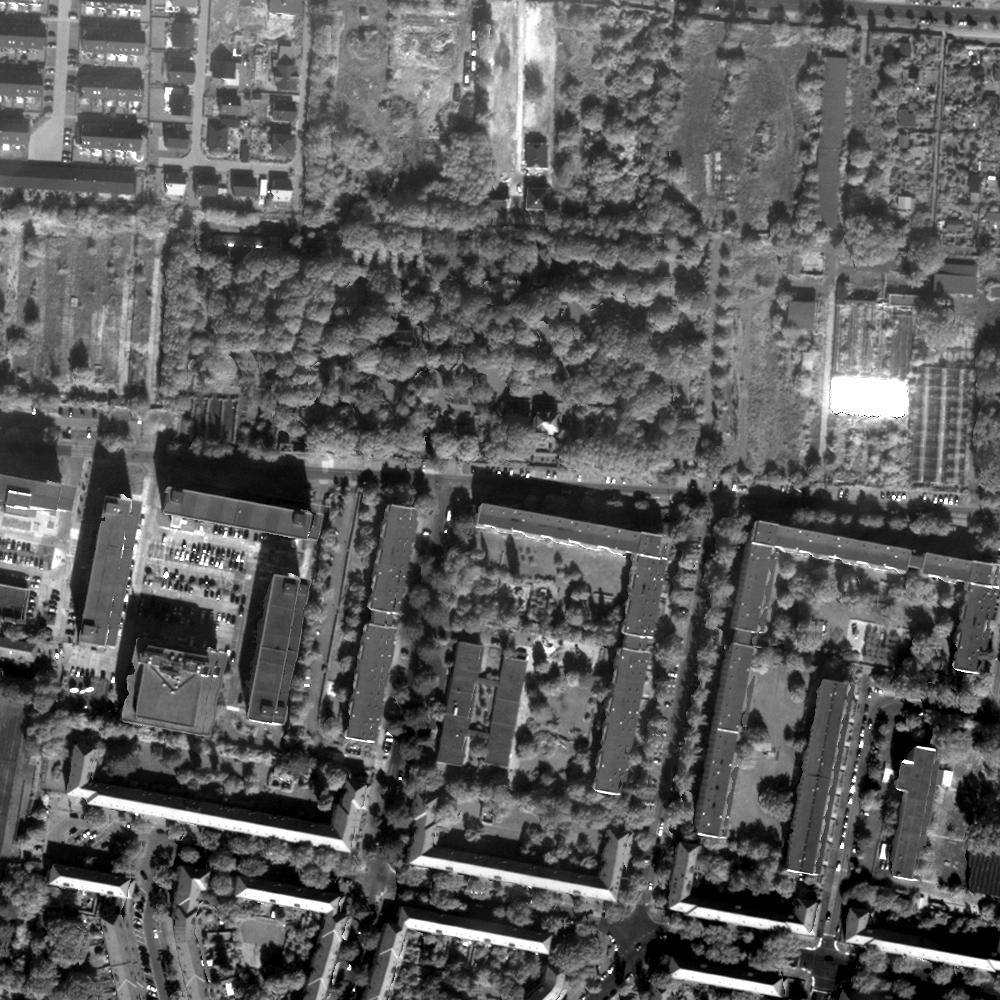}}
\subfigure[Photogrammetric DSM]{\includegraphics[width=0.3\textwidth]{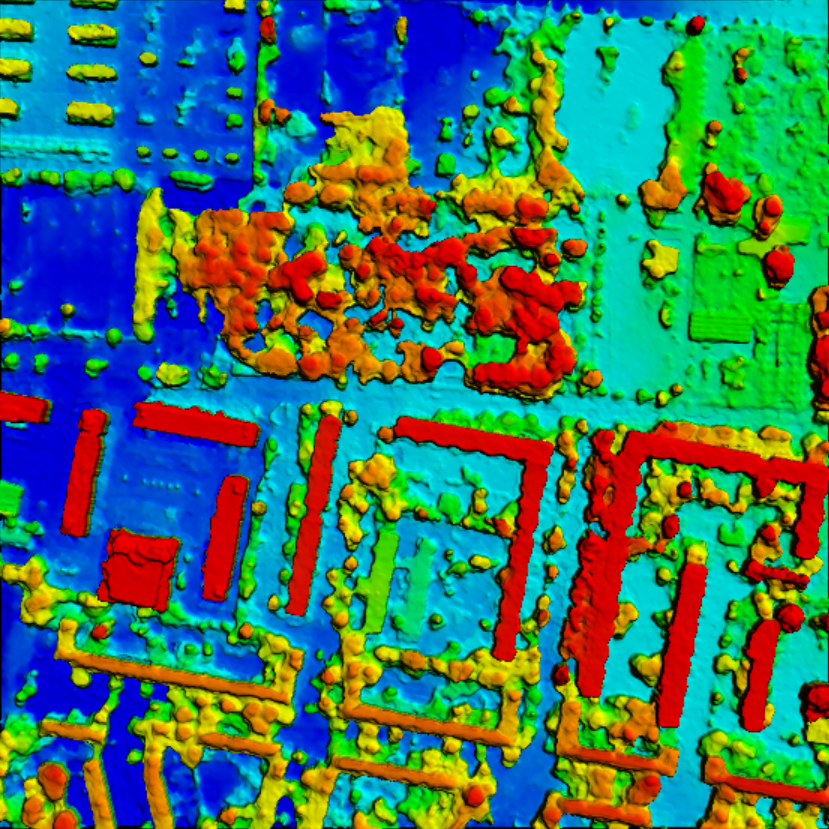}}
\subfigure[Ground truth DSM]{\includegraphics[width=0.3\textwidth]{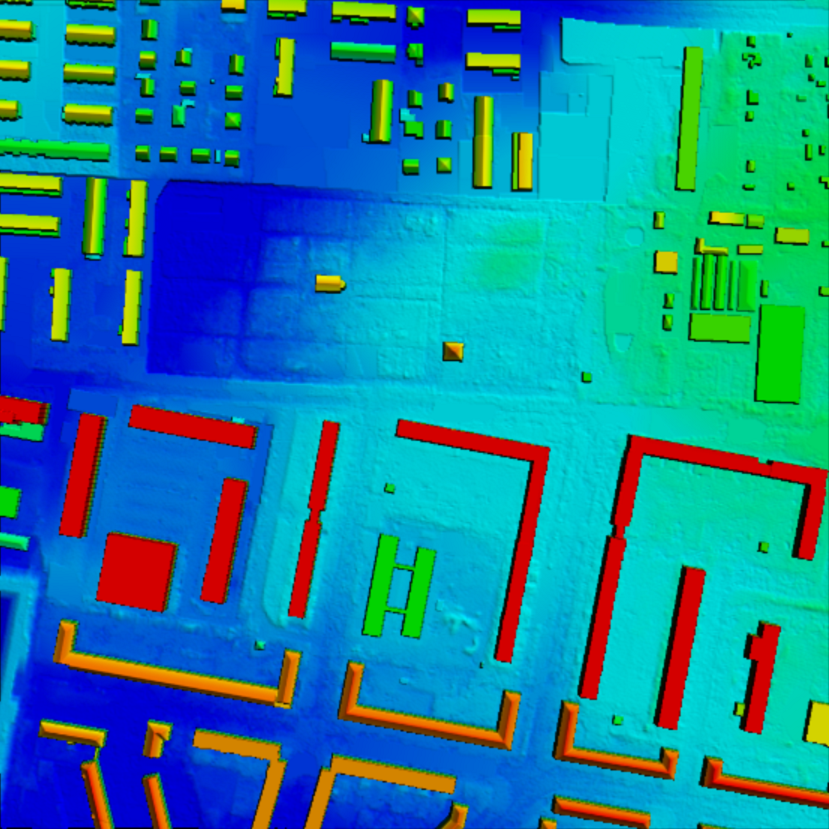}}
\subfigure[Refined DSM]{\includegraphics[width=0.3\textwidth]{Figures/results/dsm_out_qt.png}}
\subfigure[Detected edges and corners]{\includegraphics[width=0.3\textwidth]{Figures/results/edge_out_w.png}}
\subfigure[Vectorized edges and corners]{\includegraphics[width=0.3\textwidth]{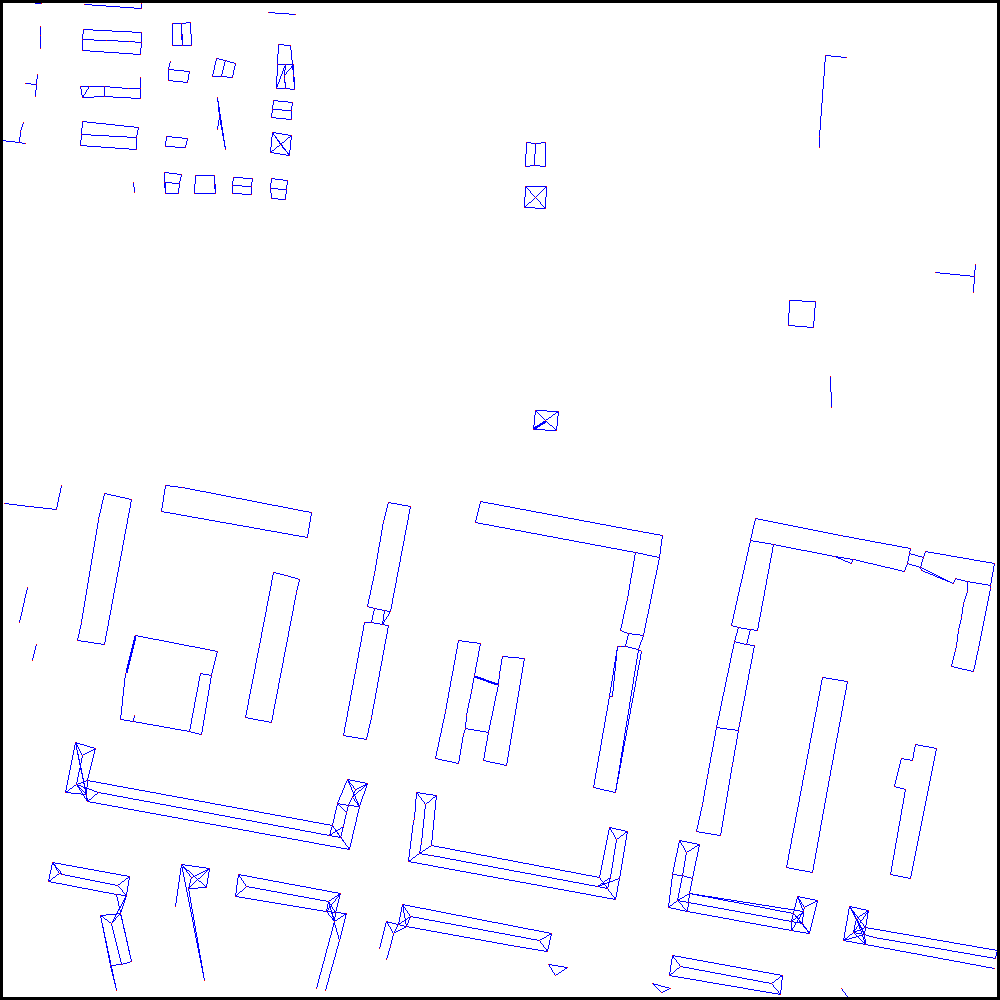}}

\caption{Experimental results of a 500m $\times$ 500m testing area. Some buildings in $(c)$ are not shown in other images because of the time difference. Some edges are missing in $(f)$ compared to $(e)$ because they don't meet the requirements of vectorization process, especially for boundary objects as they are incomplete.}
\label{fig:experiment-results}
\end{figure*}

\Cref{fig:experiment-results} $(c)$ shows the \gls{gl:DSM} refinement result, from which it can be seen that the proposed network can both filter out and regularize building objects from the photogrammetric \gls{gl:DSM}. This in parallel shows the robustness and accuracy of our approach to detect correct buildings, as we can see from \cref{fig:experiment-results} $(d)$ that the ground truth consists of several buildings that are not shown in satellite images due to the time difference. \Gls{gl:MAE}, \gls{gl:RMSE} and \gls{gl:NMAD} are applied for quantitative evaluation of the \gls{gl:DSM} refinement result: 
\begin{equation}
\label{MAE}
\varepsilon_{\mathrm{MAE}}(h, \hat{h})=\frac{1}{n} \sum_{j=1}^{n}\left|\hat{h}_{j}-h_{j}\right|
\end{equation}

\begin{equation}
\label{RMSE}
\varepsilon_{\mathrm{RMSE}}(h, \hat{h})=\sqrt{\frac{1}{n} \sum_{j=1}^{n}\left(\hat{h}_{j}-h_{j}\right)^{2}}
\end{equation}

\begin{equation}
\label{NMAD}
\varepsilon_{\mathrm{NMAD}}(h, \hat{h})=1.4826 \cdot {median}_{j}\left(\left|\Delta h_{j}-m_{\Delta h}\right|\right)
\end{equation}
\noindent
where $\hat{h}$ denotes the predicted heights, $h$ denotes the target heights, $\Delta h$ denotes height error and $m_{\Delta h}$ denotes median height error. As is shown in \cref{tab:DSM-metrics-1}, our network improves all three metrics evaluated over the testing area compared to Bittner \etal~\cite{bittner2020long}. The \glspl{gl:RMSE} of all \glspl{gl:DSM} are relatively large compared to the ground truth, which can be explained by the time difference between the reference data and the given satellite \gls{gl:DSM}. There can be cases when in one data source the buildings exist and in the other not (due to new buildings construction or their destruction), and vice versa.

\begin{table}
\caption{Quantitative metrics for refined \gls{gl:DSM} evaluated over the testing area.}
\label{tab:DSM-metrics-1}
\centering
\scalebox{0.8}{
\begin{tabular}{c c c c}
\toprule
  & \textbf{MAE (m)} & \textbf{RMSE (m)} & \textbf{NMAD (m)} \\
\midrule
Photogrammetric DSM & 3.91 & 7.14 & 1.40\\
Bittner \etal~\cite{bittner2020long} & 1.73 & 4.02 & 0.93\\
Ours (with attention) & \textbf{1.42} & \textbf{3.65} & \textbf{0.60}\\
\bottomrule
\end{tabular}
}
\end{table}


\Cref{fig:experiment-results} $(e)$ and $(f)$ present the edge and corner detection and vectorization results. By combining building height and shape information from the refined DSM and intensity information from the PAN image, the results show well-formed building skeletons with accurate corners and complete outlines. As a result of the requirements from the vectorization process, edges which have only one or none corner detected, or which are over-curved are unable to be determined. However, though missing some of the expected line segments, most of the building outer boundaries and inner edges are successfully constructed. Meanwhile, it might be helpful to mention that during the experiments we tried also combining the two steps (\gls{gl:DSM} refinement and edge and corner detection) together in a multi-task manner, but the results got worse, as the edge and corner detection network benefits more from an already refined \gls{gl:DSM} as input.

\begin{figure}
\centering
\includegraphics[width=\linewidth]{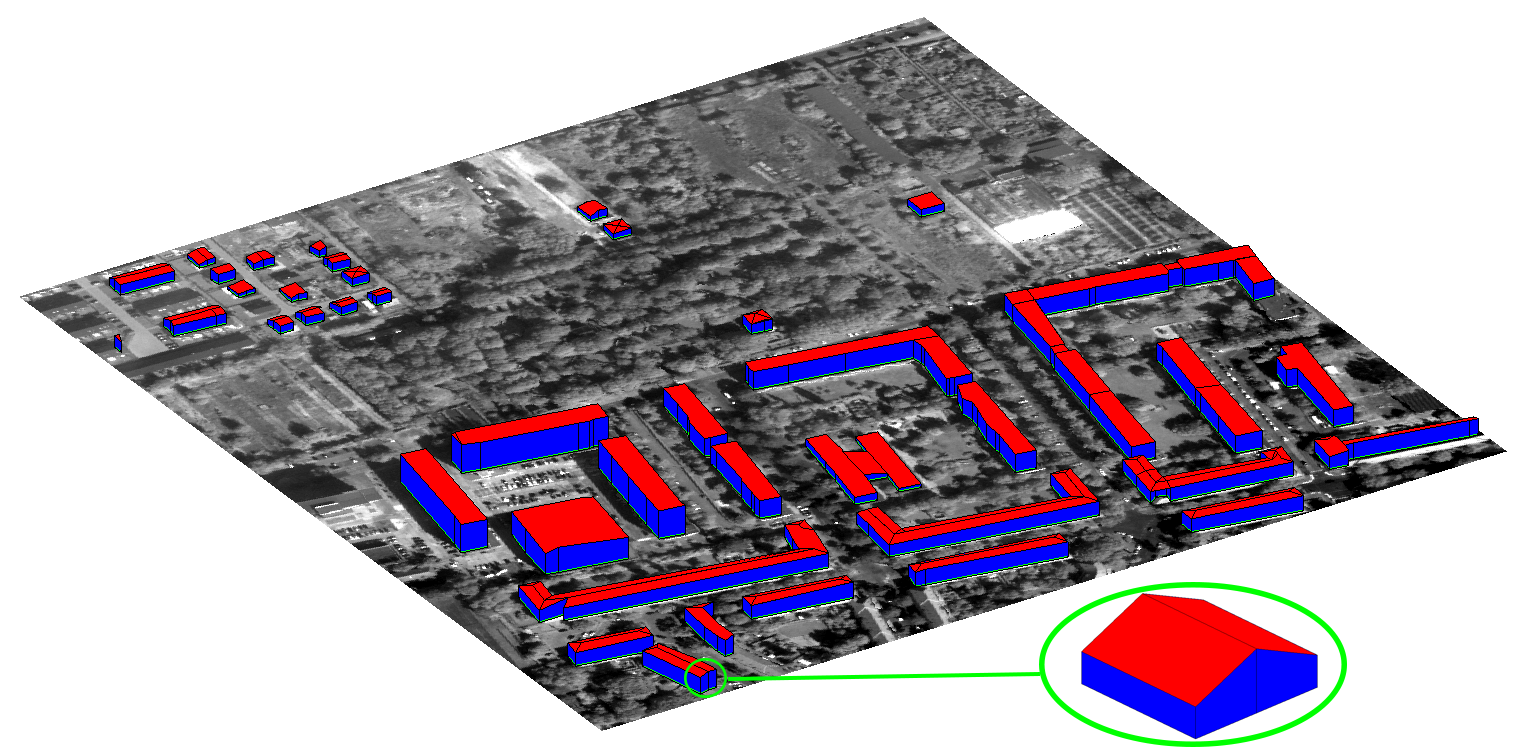}
\caption{Reconstructed 3D building model of a $500m\times 500m$ testing area.}
\label{fig:result-3Dmodel}
\end{figure}

The final vectorized 3D building model is shown in \cref{fig:result-3Dmodel}, where most of the buildings are well reconstructed in 3D space. Even though some buildings are not fully visible in PAN image and blurry in photogrammetric \gls{gl:DSM}, we can still reconstruct them to a good shape. It is also seen that some buildings are missing or incomplete, which is due to the missing of those vectorized edges and corners whose quality doesn't meet the vectorization process.

For quantitative evaluation of the height of reconstructed buildings, the generated \gls{gl:nDSM} is compared to the ground truth. \gls{gl:MAE}, \gls{gl:RMSE} and \gls{gl:NMAD} are applied again to evaluate the quality of the generated \gls{gl:nDSM}. The evaluation result is shown in \cref{tab:nDSM-metrics}, from which we can see that both photogrammetric nDSM and our genereated \gls{gl:nDSM} have better metrics than \glspl{gl:DSM} (\cref{tab:DSM-metrics-1}) after removing the height of ground surface. Meanwhile, our result presents large improvement compared to photogrammetric \glspl{gl:nDSM}.

\begin{table}
\caption{Quantitative metrics for building \gls{gl:nDSM} evaluated over the testing area.}
\label{tab:nDSM-metrics}
\centering
\scalebox{0.8}{
\begin{tabular}{c c c c}
\toprule
 & \textbf{MAE (m)} & \textbf{RMSE (m)} & \textbf{NMAD (m)} \\
\midrule
Photogrammetric nDSM & 3.21 & 6.04 & 0.85 \\
Ours & \textbf{0.80} & \textbf{2.28} & \textbf{0.47} \\
\bottomrule
\end{tabular}
}
\end{table}

\begin{table}
\caption{Quantitative metrics for roof orientation error evaluated over the testing area.}
\label{tab:orientation-metrics}
\centering
\scalebox{0.8}{
\begin{tabular}{c c c c c}
\toprule
 & \textbf{min ($^{\circ}$)} & \textbf{max ($^{\circ}$)} & \textbf{mean ($^{\circ}$)} & \textbf{$\sigma$ ($^{\circ}$)} \\
\midrule
Photogrammetric nDSM & \textbf{0.08} & 75.84 & 22.46 & 22.28 \\
Ours & 0.10 & \textbf{75.83} & \textbf{9.31} & \textbf{15.53} \\
\bottomrule
\end{tabular}
}
\end{table}

To evaluate the quality of the reconstructed 3D roofs, an orientation error is applied to examine the inclination of the constructed roof planes. As proposed by Koch \etal~\cite{koch2018evaluation}, the orientation error can be formulated as the angle difference between the normal vectors of 3D planes fitted to the predicted surface points and the given ground truth points:

\begin{equation}
\varepsilon_{\text {orie }}\left(G \odot \mathcal{P}\right)=\arccos \left(n_{i}^{\mathrm{t}} \cdot \check{n}_{i}^{\mathrm{p}}\right)
\end{equation}

\noindent
where $n_{i}^{\mathrm{t}}$ and ${n}_{i}^{\mathrm{p}}$ denote the normal vector of a certain plane on target and predicted image respectively. $G\odot \mathcal{P}$ represents the predicted depth image $G$ masked with a binary mask $\mathcal{P}$ containing a certain number of roof planes. \Cref{tab:orientation-metrics} shows the average orientation error of constructed 3D roof faces compared to corresponding ground truth, showing that the average plane angle is within 10$^{\circ}$, which is much better than using only the photogrammetric \gls{gl:nDSM}.

\begin{figure*}[ht]
\centering

\subfigure[PAN image]{
\includegraphics[width=0.35\textwidth]{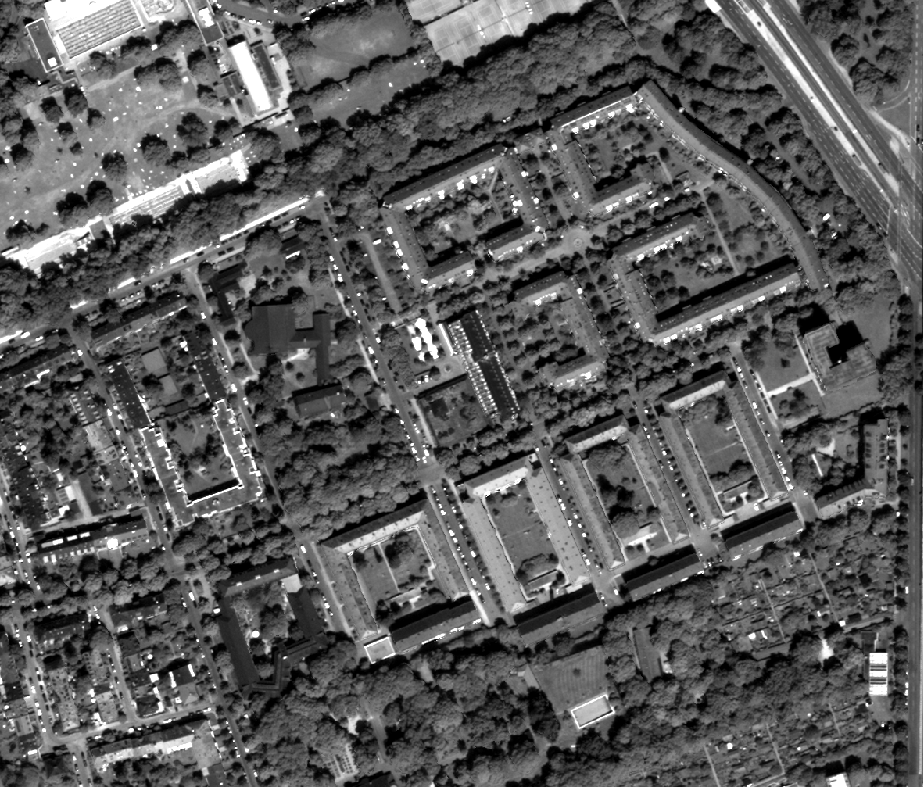}
}
\subfigure[Photogrammetric DSM]{
\includegraphics[width=0.35\textwidth]{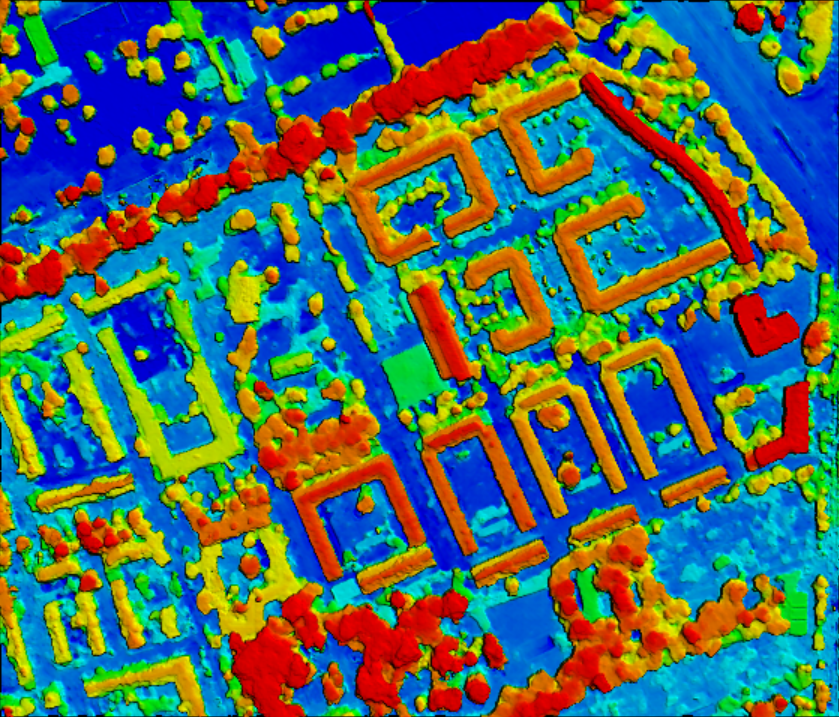}
}
\subfigure[Refined DSM]{
\includegraphics[width=0.35\textwidth]{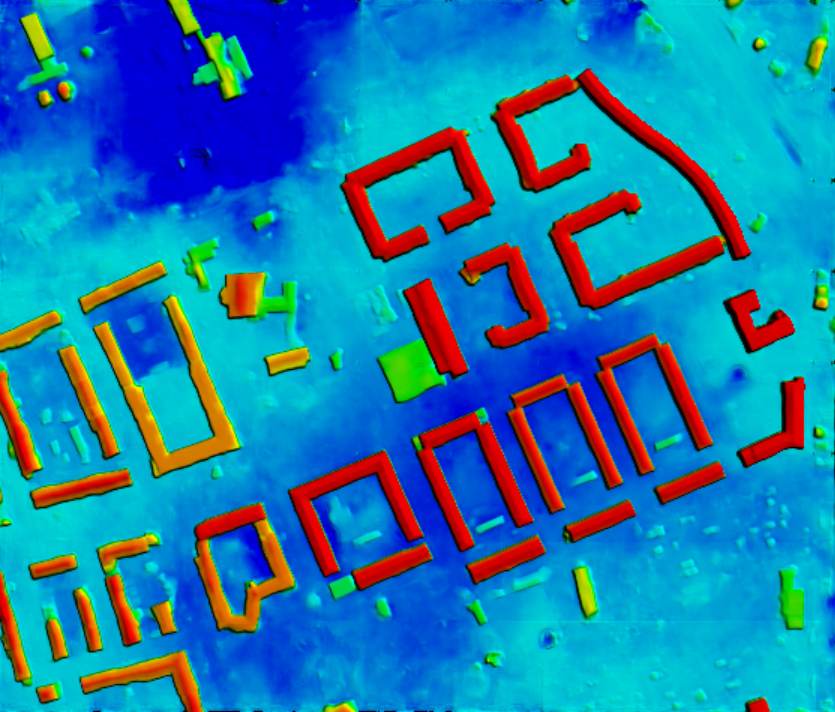}
}
\subfigure[Reconstructed 3D model]{
\includegraphics[width=0.35\textwidth]{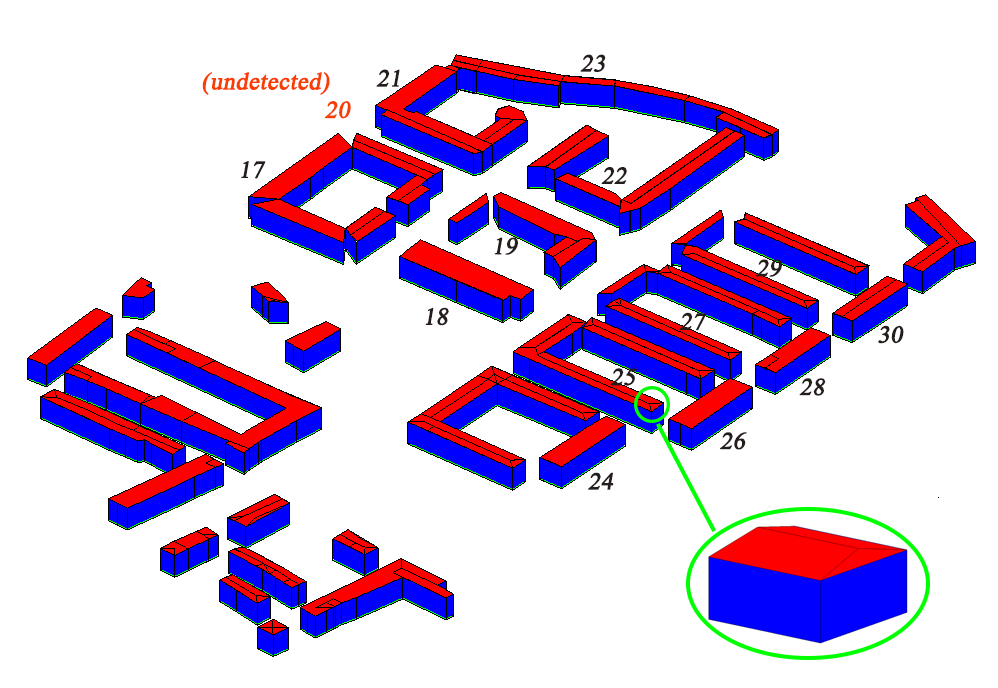}
}

\caption{Testing results of a sub-area of Munich.}
\label{fig:result-munich}
\end{figure*}

\begin{table}
\caption{Comparison of eave and ridge heights of the building model for selected buildings.}
\label{tab:ridgeandeave}
\centering
\scalebox{0.55}{
\begin{tabular}{| c | c | c | c | c | c | c |}
\hline
\multirow{2}{*}{Building No.} & \multicolumn{3}{c|}{Ridge (m)}  & \multicolumn{3}{c|}{Eave (m)} \\
\cline{2-7}
& Reference & Partovi \etal \cite{partovi2019automatic} & Ours & Reference & Partovi \etal \cite{partovi2019automatic} & Ours \\
\hline
17 & 15   & 14.03 & 15.05 & 11   & 11.29 & 11.72 \\
\hline
18 & 19   & 17.46 & 18.21 & 15   & 13.38 & 15.50 \\
\hline
19 & 15   & 14.42 & 16.13 & 11   & 12.52 & 13.01 \\
\hline
20 & 15   & 14.22 & - & 11   & 10.86 & - \\
\hline
21 & 15.5 & 14.08 & 15.33 & 11.9 & 12.21 & 11.54 \\
\hline
22 & 15.6 & 15.28 & 14.87 & 11.5 & 11.87 & 11.94 \\
\hline
23 & 20.0 & 20.76 & 21.80 & 16.5 & 17.35 & 17.58 \\
\hline
24 & 16.2 & 15.87 & 17.03 & 12.3 & 11.03 & 13.66 \\
\hline
25 & 17.4 & 16.21 & 18.02 & 13.6 & 13.58 & 13.77 \\
\hline
26 & 16.8 & 16.40 & 17.19 & 12.5 & 10.54 & 11.36 \\
\hline
27 & 15   & 14.66 & 13.88 & 10.9 & 10.49 & 11.40 \\
\hline
28 & 16.8 & 16.27 & 17.11 & 12.5 & 10.41 & 12.96 \\
\hline
29 & 14.7 & 13.94 & 15.54 & 10.7 & 10.46 & 11.06 \\
\hline
30 & 16.8 & 15.66 & 16.00 & 12.5 & 10.51 & 13.79 \\
\hline \hline
$\mu_{|\Delta H|}$ & - & 0.79 & \textbf{0.74} & - & \textbf{0.93} & 0.99 \\
\hline
$\sigma_{|\Delta H|}$ & - & \textbf{0.41} & 0.44 & - & 0.77 & \textbf{0.74} \\
\hline
RMSE & - & 0.89 & \textbf{0.74} & - & \textbf{1.20} & 1.54 \\
\hline
NMAD & - & \textbf{0.55} & 0.59 & - & 0.82 & \textbf{0.68} \\
\hline

\hline 
\end{tabular}
}
\end{table}

In addition, we compare our proposed 3D building vectorization method with the work presented by Partovi \etal~\cite{partovi2019automatic}, who developed a multi-stage hybrid method for 3D building reconstruction using \gls{gl:PAN} images, photogrammetric \glspl{gl:DSM} and multi-spectral images from satellite data. \cref{fig:result-munich} presents the reconstruction results of a sub-area of Munich using Worldview-2 satellite data. The ridge and eave heights of 14 reconstructed buildings in this area are compared with reference data from the Department of Urban Planning and Building (DUPB) of Munich. As is shown in \cref{tab:ridgeandeave}, $|\Delta H|$ denotes the absolute height difference between the predicted model and reference, and $\mu_{|\Delta H|}$ and $\sigma_{|\Delta H|}$ represent the mean and standard deviation of the height difference, respectively. The building numbers refer to \cref{fig:result-munich} $(c)$. It can be seen that both methods lead to lower accuracy in eave heights than ridge heights because the surroundings of building boundaries are usually more complex than inner-roof ridges both in PAN image and photogrammetric \gls{gl:DSM}. Our method tends to get bigger values for eave heights, which can be explained by our valuing method for the height of building corners. In order to avoid the mismatching between \gls{gl:DSM} heights and corner positions, we give the eave corner the maximum height value in a surrounding window and the minimum height value for the corresponding ground corner. This would increase the relative height of the building eaves, yet we can see that this systematic error is within a small range. Apart from that, the overall accuracy shows promising superiority of our method, where we get comparative metric performance with a simpler approach. Meanwhile, as a price of simplicity, the biggest problem remaining to be solved is the lack of completeness of our constructed model. As can be seen from both \cref{fig:result-3Dmodel} and \cref{fig:result-munich}, some building components are lost after vectorization, which quantitatively reduces the recall score from $0.88$ to $0.81$ (Berlin testing area) compared to the refined \gls{gl:DSM} before vectorization.

\glsresetall
\section{Conclusion}

In this paper, we present a multi-stage large-scale 3D building vectorization approach. We extend the application of recent deep learning based techniques on photogrammetric \gls{gl:DSM} refinement and bring it to the application of automatic 3D building model reconstruction. With the help of a self-attention module, we obtain promising results for both regression of building heights and semantic segmentation of edges and corners. Based on that, we propose a simple yet effective vectorization pipeline to reconstruct \gls{gl:LoD}-2 building models. We apply \gls{gl:NMS} to filter out best fitting corner points, define buffer connectivity and buffer thresholds to determine edges, and polygonize them to roof faces. By utilizing again the height information from the refined \gls{gl:DSM}, we finally reconstruct fully vectorized 3D building models. Though limitations exist in straight edge assumptions and the completeness of reconstructed building models, results prove the overall robustness and accuracy of our proposed method.

{\small
\bibliographystyle{ieee_fullname}
\bibliography{egbib}
}

\end{document}